\documentclass{article}
\usepackage{microtype}
\usepackage{graphicx}
\usepackage{subcaption}
\usepackage{booktabs}

\usepackage{hyperref}

\usepackage[preprint]{icml2026}

\usepackage{amsmath}
\usepackage{amssymb}
\usepackage{mathtools}
\usepackage{amsthm}

\usepackage[capitalize,noabbrev]{cleveref}

\theoremstyle{plain}

\theoremstyle{definition}

\theoremstyle{remark}

\usepackage[textsize=tiny]{todonotes}

\icmltitlerunning{}

\begin{document}

\twocolumn[
\icmltitle{LLM Evaluation on Unseen Questions: Contextual Multidimensional IRT Model}

  \icmlsetsymbol{equal}{*}

  \begin{icmlauthorlist}
    \icmlauthor{Ergan Shang}{CMU}
    \icmlauthor{Weijing Tang}{CMU}
    \icmlauthor{Yinqiu He}{UWM}
    
  \end{icmlauthorlist}

  \icmlaffiliation{CMU}{Department of Statistics and Data Science, Carnegie Mellon University}
  \icmlaffiliation{UWM}{Department of Statistics, University of Madison}

  \icmlcorrespondingauthor{Weijing Tang}{weijingt@andrew.cmu.edu}
  \icmlcorrespondingauthor{Yinqiu He}{yinqiu.he@wisc.edu}

  \icmlkeywords{LLM evaluation, out-of-sample performance prediction, multidimensional item response theory, contextual embeddings}

  \vskip 0.3in
]

\printAffiliationsAndNotice{}

\begin{abstract}

Evaluation of large language models (LLMs) increasingly requires predicting 
  how a model will perform on new questions or tasks  before collecting large amounts of new annotations. This problem is challenging because question difficulty, scenario, and underlying capability demands can vary substantially.  Simple retrospective averages may confound model ability with item characteristics. In this paper, we study a model-based evaluation framework that combines multidimensional item response theory model  with question contexts to predict LLM performance on unseen questions. The framework represents LLMs through latent capability profiles while using question content to inform item characteristics, allowing information to transfer beyond previously observed items. Empirically, we find that for within-scenario evaluation, incorporating question embeddings improves prediction relative to model-free baselines, and that multidimensional latent structure provides a richer description of capability variation than unidimensional alternatives. At the same time, our results reveal an important limitation that the generalizability does not necessarily translate into reliable prediction under cross-scenario shift.
  These findings suggest that context-aware psychometric modeling is a promising direction for efficient and interpretable LLM evaluation, while also highlighting cross-scenario generalization as a central open challenge.
\end{abstract}

\vspace{-2em}

\section{Introduction}

Evaluation of large language models (LLMs) is increasingly shaped by rapid model iteration, evolving benchmarks, and deployment settings that change faster than full-scale annotation pipelines can keep up \citep{liang2023holistic,you2024llm,white2025livebench,shang2026eraseearlybackpropagationschedule}. In this setting, a central question is no longer only how a model performed on a fixed benchmark, but how well observed evaluation results can predict behavior on new questions. Reliable prediction on unseen questions is crucial for deployment decisions,  because it helps distinguish genuine generalization from benchmark-specific tuning and supports more trustworthy model selection and risk assessment \cite{truong2025reliable,li2025activeEval}.
This is especially important in high-stakes or rapidly changing applications, where collecting new gold-standard labels may require costly human annotators or expensive judge models \citep{pacchiardi2024hundredInstances}.
In such cases, leveraging existing or small-scale evaluations to predict performance on new questions can substantially improve evaluation efficiency and guide resource allocation
\cite{polo2024tinyBenchmarks,wang2025essencebench}.

Reliable prediction is challenging, however, because benchmark items are heterogeneous. Questions can vary substantially in difficulty,  context, and the underlying capabilities they demand \citep{zhuang2025psychometricAI}. 
As a result, simply averaging correctness across questions can confound model capabilities with question characteristics, obscuring   structures that matter for extrapolation to unseen questions  \cite{lalor2016evaluationScale}. 

Item response theory (IRT) serves as  a natural approach to address this problem \citep{baker2001basics,chen2025item}. 
Prior studies have used IRT to build evaluation scales, estimate latent item parameters from model response patterns, analyze benchmark quality, and enable more reliable or amortized evaluation \cite{lalor2016evaluationScale,lalor2019artificialCrowds,lalor2024irtNLP,polo2024tinyBenchmarks}.
More recently, several works have combined IRT modeling with question embeddings to estimate item characteristics and support prediction on previously unseen questions  \cite{mccarthy2021jumpStarting,marinho2023predictingIRTtext,truong2025reliable,tang2026knowledge}.
However, most existing approaches remain largely based on unidimensional ability, which may not capture the multiple latent capabilities that drive LLM performance. They also leave open a harder question: \emph{whether predictive evaluation can generalize across heterogeneous scenarios} rather than only to held-out questions within the same benchmark setting.
This broader scope of generalization is especially important when evaluating LLMs on new tasks or benchmarks \citep{phan2025humanity,li2026long,zhou2026asibenchdawnartificialsuperintelligence}. 

In this work, we study LLM evaluation on unseen questions through a contextual multidimensional IRT (MIRT) model. Our model jointly captures question characteristics and model capabilities, allowing semantic information from question text to inform item parameters while providing a richer description of latent capability structure.

Our main findings are threefold. First, incorporating question embeddings into MIRT models improves predictive performance relative to model-free alternatives. Second, allowing multiple latent dimensions yields a richer description of model capability than a purely unidimensional approach. Third, although the proposed framework performs encouragingly when predicting held-out questions within the same scenario, its cross-scenario performance has larger variation, suggesting that transfer learning across domains remains a major open challenge. In summary, these results show context-aware psychometric modeling is a promising direction for efficient and interpretable LLM evaluation, while cautioning that within-scenario predictive success may not necessarily translate to   robust cross-scenario generalization.

\subsection{Related work}

Benchmark-based evaluation remains   dominant   for comparing LLMs, but its limitations in both generalizability and efficiency are increasingly recognized.  Prior studies have argued that conclusions drawn from static benchmarks may not directly transfer to changing evaluation protocols  \citep{alzahrani2024benchmarks} or scenarios 
\cite{kiela2021dynabench,lin2025wildbench}.
Meanwhile, a growing line of work seeks   to reduce evaluation cost by recovering full-benchmark conclusions from only a small subset of instances
\cite{polo2024tinyBenchmarks,pacchiardi2024hundredInstances,li2025activeEval,wang2025essencebench,zhong2025collaborativeFiltering}.
Our work shares the goal of moving beyond  retrospective scoring on a fixed benchmark, but studies a structured model-based approach for predicting performance  on unseen questions.

 Achieving this goal requires separating model-side abilities from heterogeneous  question-side characteristics. 
IRT has long offered a principled way to disentangle item properties from participant abilities in psychometrics \cite{baker2001basics,cai2016item}. In natural language processing (NLP), \citet{lalor2016evaluationScale,lalor2019artificialCrowds} introduce  IRT-based evaluation scales.
The recent tutorial by \citet{lalor2024irtNLP} highlights growing interest in IRT as a general framework for   model assessment in language technology. In the LLM setting, an emerging number of studies have used IRT-based modeling to study benchmark measurement     quality or interpretations \cite{zhou2025lostBenchmarks,yao2025jeirt,cai2025latent}.

A parallel line of work  leverages question texts or embedding-based features to model item characteristics and support generalization to unseen items. In educational assessment, such approaches have been used for predicting  difficulty \citep{alkhuzaey2024questionDifficulty,marinho2023predictingIRTtext}  or unseen  items   \citep{mccarthy2021jumpStarting,khan2025justReadQuestion}. For LLM evaluation, the work most closely related to ours is \citet{truong2025reliable}, who combine Rasch-style psychometric modeling  and question embeddings to support reliable and efficient amortized evaluation.

Our method builds on this line of work but differs in two key respects. First, we explicitly model multivariate  latent capability dimensions of LLMs rather than relying on a unidimensional Rasch-model view or hard-to-interpret blackbox predictors. Second, we focus on the generalizability of evaluation prediction on unseen questions, including the more difficult setting of cross-scenario transfer rather than only efficient estimation on a fixed benchmark.

\section{Method} 
To predict evaluations of LLMs on unseen questions, we use the following  contextual multidimensional item response theory (C-MIRT) model that separates model-side latent abilities from question-side characteristics across scenarios.

\paragraph{C-MIRT model.}

Consider a benchmark dataset where models are evaluated across $S$  different scenarios, i.e.,  different task types or contextual domains. For each scenario $s\in [S]$, suppose $n $ LLMs are evaluated on $p_s$ questions.
Let $y_{ij}^{(s)}\in \{0,1\}$ denote whether model \(i\) answers question \(j\) correctly in scenario \(s\). Each question $j$ is associated with a contextual embedding $e_j\in \mathbb{R}^d$. To model performance on these questions, we map this embedding into an \(r\)-dimensional latent question representation through a feature map $\phi^{(s)}:\mathbb{R}^d\to\mathbb{R}^r$. Then, the response probability in the C-MIRT is modeled through a logistic link as 
\[
y_{ij}^{(s)}\mid \theta_{ij}^{(s)} \sim \mathrm{Bernoulli}\!\left(\sigma(\theta_{ij}^{(s)})\right),
\qquad
\sigma(x)=\frac{1}{1+e^{-x}},
\]
with the parameter
\begin{align}\label{eq:bilinear}
    \theta_{ij}^{(s)} =  {\alpha_i^{(s)}} + { {u_i^{(s)}}}^\top \phi^{(s)}(e_{j}).
\end{align}
Here \(\alpha_i^{(s)}\in\mathbb{R}\) is a scenario-specific intercept for model~\(i\), capturing its baseline success rate in scenario \(s\), and \(u_i^{(s)}\in\mathbb{R}^r\) is a scenario-specific latent capability vector. 
The transformed embedding \(\phi^{(s)}(e_j)\) represents the latent characteristics of question \(j\) in scenario \(s\). 
The bilinear form $u_i^{(s) \top} \phi^{(s)}\left(e_j\right)$ captures how well the capabilities of model \(i\) align with the requirements of question \(j\). 
Different specifications of \(\phi^{(s)}\) can be used to incorporate contextual information about the questions.
For example, \citet{tang2026knowledge} models \(\phi^{(s)}\) as a smooth function in a reproducing kernel Hilbert space, so that questions with similar contextual embeddings are encouraged to have similar latent representations.

The C-MIRT formulation goes beyond a unidimensional difficulty-based view by allowing model performance to vary along multiple latent traits. 
In particular, the C-MIRT model is closely related to the Rasch-style formulation in \citet{truong2025reliable}, which models
\[
\theta_{ij}^{(s)}=\alpha_i^{(s)}-\beta_j^{(s)},
\]
with a scalar difficulty parameter \(\beta_j^{(s)}\) for question \(j\) in scenario \(s\). 
In that formulation, all item heterogeneity is absorbed into a single difficulty dimension. In contrast, C-MIRT in \eqref{eq:bilinear} takes a multivariate bilinear form, which allows question characteristics and model capabilities to interact through multiple latent dimensions.

An additional advantage of C-MIRT is that it naturally supports prediction on unseen questions. Once the feature map \(\phi^{(s)}\) is learned from training questions in scenario \(s\), a new question can be embedded into the same latent space and used to predict response probabilities. This allows us to investigate the generalization within and across scenarios.

\paragraph{Model estimation.}
Given a source scenario \(s\), we estimate the model parameters \(({\alpha}_i^{(s)}, {u}_i^{(s)}, {\phi}^{(s)})\) by minimizing the logistic negative log-likelihood under the C-MIRT model, where the feature map ${\phi}^{(s)}$ is parameterized by a multilayer perceptron. Because the parameters in MIRT are only identifiable up to a linear transformation, we adopt a two-step estimation procedure; details are deferred to the Appendix. 

\paragraph{Predicting responses to unseen questions.}
Suppose \(\widehat{\alpha}_i^{(s)}, \widehat{u}_i^{(s)}, \widehat{\phi}^{(s)}\) are estimated from the training split of the source scenario \(s\). For a question \(j\) from target scenario \(t\), with embedding \(e_{j}\), we compute 
\[
\widehat{\theta}_{ij}^{(s\rightarrow t)}
=
\widehat{\alpha}_i^{(s)}
+
\widehat{u}_i^{(s)\top}\widehat{\phi}^{(s)}(e_{j}),
\qquad
\widehat{p}_{ij}^{(s\rightarrow t)}
=
\sigma\!\left(\widehat{\theta}_{ij}^{(s\rightarrow t)}\right).
\]
When \(t=s\), this gives within-scenario predictions for previously unseen questions, enabled by the learned feature map \(\widehat{\phi}^{(s)}\). When \(t\neq s\), this gives cross-scenario predictions. This directly evaluates out-of-scenario generalization, and the performance depends on how well the latent structure learned under scenario \(s\) transfers to scenario \(t\), as we demonstrate empirically in Section~\ref{sec:cmirt_pred}.

\section{Experiments}
We adopt the benchmark setup  in \citet{truong2025reliable}, which leveraged 22 datasets from 5 HELM repositories: Classic, Lite, AIR-Bench, Thai Exam, and MMLU. We filtered out  scenarios where the question descriptions are incomplete,  e.g., questions with missing multiple-choice options. After filtering, we retain 11 scenarios for evaluation. Examples of scenarios include ``wikifact" and ``math". The number of questions per scenario ranges from 436 to 29407. 

We construct a contextual embedding $e_j$ for each benchmarking question using BERT-based language models (\citet{wang2020minilm,reimers2019sentence}). 

All methods are evaluated using 5-fold cross-validation. Within each scenario, questions are split into five folds. In  each run, one fold, containing 20\% of questions, is used for testing, and the remaining four folds,   containing 80\% of questions, are used for training. 

For each training-test scenario pair and each fold, the fitted model produces a predicted score matrix, 
where rows correspond to $n$ LLMs and columns correspond to $p_{s,test}$ testing questions in  scenario $s$. We evaluate this matrix in two  ways. The first is \textbf{question-wise AUC}, or column-wise AUC, where  
for each fixed test question, we compute the AUC using the predicted scores across $n$ LLMs. This reflects how well the predicted scores rank LLMs on the same question. Repeating this over $p_{s,test}$ questions yields a distribution of question-wise AUC values.  
The second is \textbf{LLM-wise AUC}, or row-wise AUC, where for a fixed LLM,
we compute the AUC using the predicted scores across $p_{s,test}$ questions. This measures how well the predicted scores rank questions for that LLM. 
Repeating this over $n$ LLMs yields a distribution of LLM-wise AUC values. 
To summarize either distribution, we report its ${90}$th percentile.

\subsection{Within- and cross-scenario prediction by C-MIRT}\label{sec:cmirt_pred}

\begin{figure}[!htbp]
    \centering
    \includegraphics[width=\linewidth]{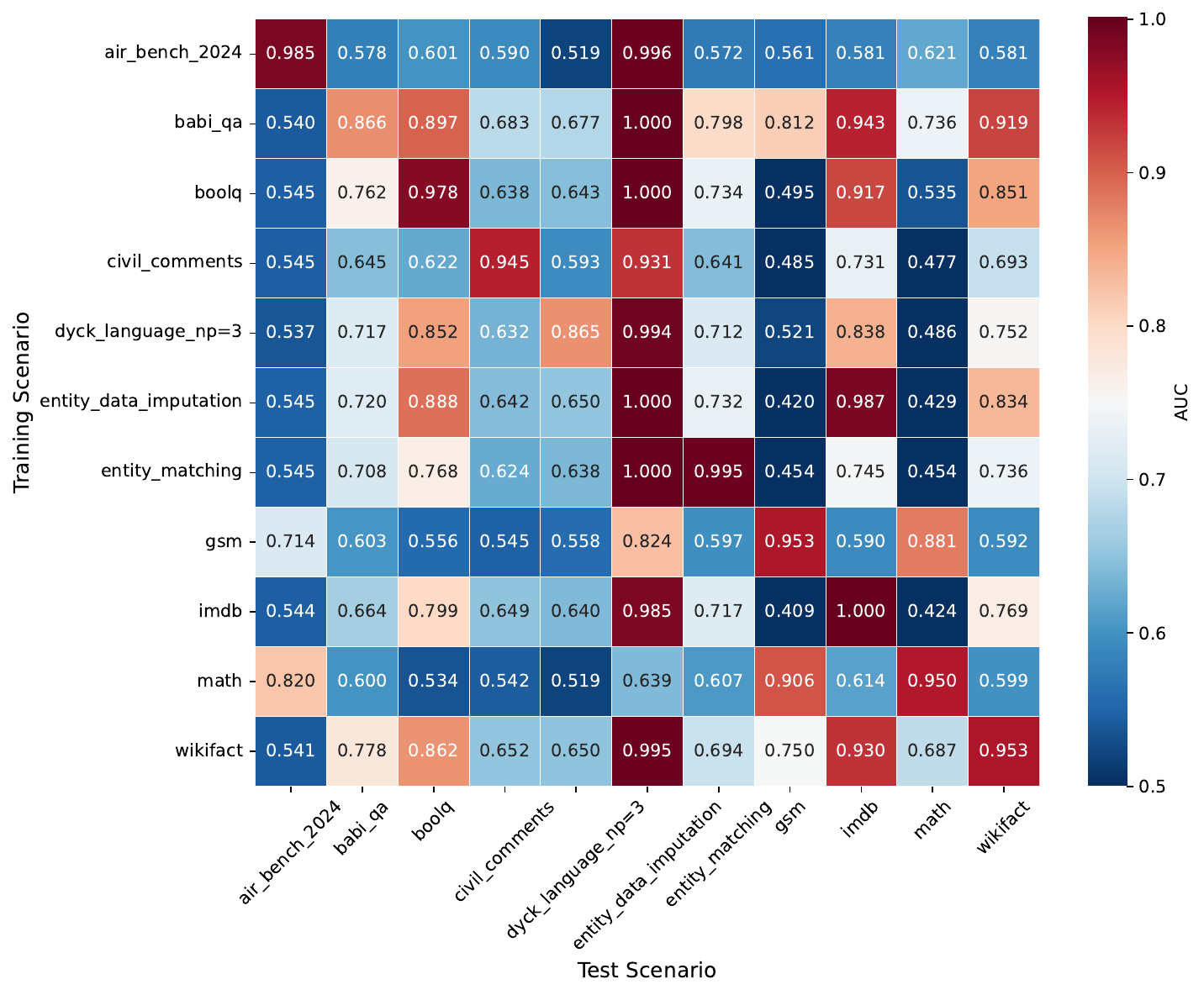}
    \caption{Heatmap of  the 90th percentile of  question-wise AUC distribution predicted by the C-MIRT method across scenarios.} 
    \label{fig:heat_col}
\end{figure}

\begin{figure}[!htbp]
    \centering
    \includegraphics[width=\linewidth]{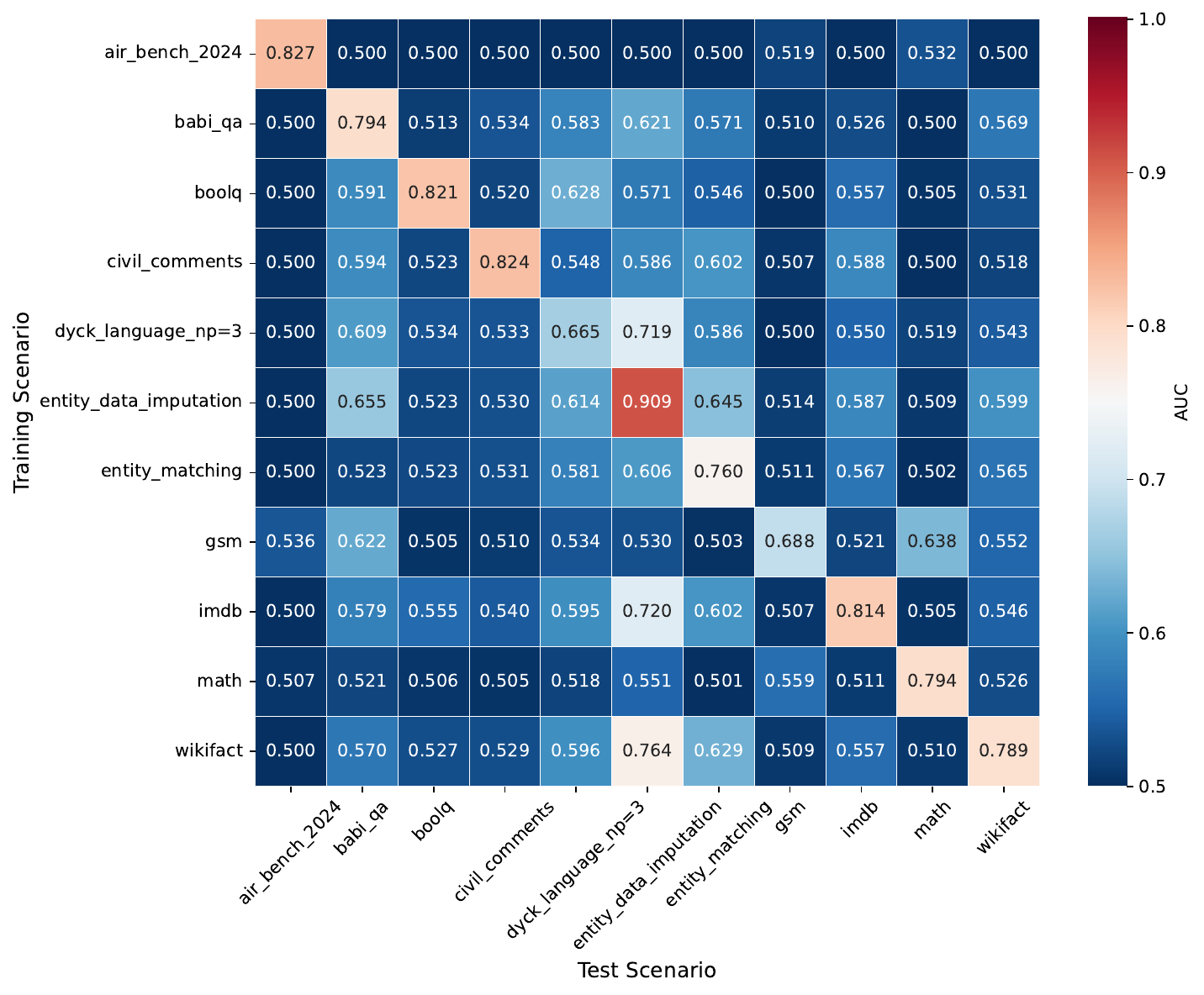}
    \caption{Heatmap of the 90th percentile of LLM-wise AUC    distribution predicted by the    C-MIRT method across scenarios. }
    \label{fig:heat_row}
\end{figure}

This section   evaluates the prediction performance of the C-MIRT.  
Figures \ref{fig:heat_col} and \ref{fig:heat_row} report the   mean 90th percentile AUC over the five cross-validation folds. In both figures, rows indicate the training scenario and columns indicate the test scenario. Diagonal entries therefore represent within-scenario prediction, whereas off-diagonal entries represent cross-scenario prediction.

Figure \ref{fig:heat_col} shows the results for question-wise AUC. The diagonal entries are consistently high, indicating strong within-scenario performance in ranking LLMs for fixed questions. Several off-diagonal entries are also relatively large, suggesting that the learned contextual structure can transfer across scenarios to some extent. For example, training scenarios such as \textit{babi\_qa}  and \textit{wikifact}, achieve comparatively strong performance on multiple test scenarios.  
Meanwhile, the substantial variation among off-diagonal cells indicates that cross-scenario transfer depends on the similarity between the source and target scenarios.

Figure \ref{fig:heat_row} shows the results for LLM-wise AUC. Compared with Figure \ref{fig:heat_col},  the diagonal entries remain strong, although the off-diagonal entries are much closer to 0.5. This suggests that ranking question difficulty for a fixed LLM is more scenario-specific and less transferable across scenarios than ranking LLMs for a fixed question. Overall, C-MIRT model performs well for within-scenario prediction in both settings, while cross-scenario generalization is notably stronger for question-wise AUC than for LLM-wise AUC.

\subsection{Comparison with competing methods} 

We compare C-MIRT  with three baselines. The first is the contextual Rasch-style model  in \citet{truong2025reliable}. The second is a lasso-penalized logistic regression model with parameters $\theta_{ij}^{(s)} = \gamma_{i0}^{(s)} + {\gamma_i^{(s)}}^\top e_{j}.$  This baseline also uses contextual embeddings $e_j$, but unlike C-MIRT, does not impose a low-rank latent structure in the coefficient matrix. 
The third is a simple mean baseline: within each scenario, the predicted score for every test question is set to the average training-set accuracy. Because this baseline assigns the same score to all questions for a given LLM, we use it only for question-wise AUC.

For each method, we compute the 90th percentile AUC in the same way as in Section~\ref{sec:cmirt_pred}. For each pair of methods, 5-fold cross-validation produces 5 paired differences between  C-MIRT and the comparator. We assess the significance of these differences using paired $t$-tests.

\begin{figure}[!htbp]
        \centering
        \includegraphics[width=\linewidth]{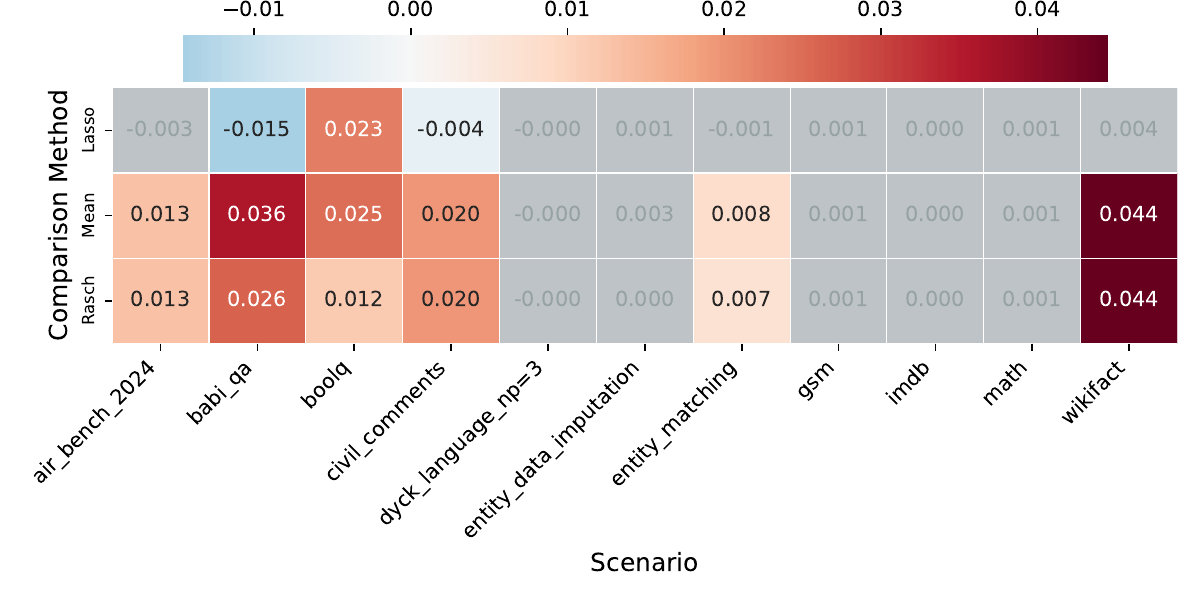}
        \caption{Heatmap of mean  differences in the 90th percentile of question-wise AUC     between     C-MIRT   and competing methods 
        across scenarios. } 
        \label{fig:compare_col}
    \end{figure}

\begin{figure}[!htbp]
        \centering
        \includegraphics[width=\linewidth]{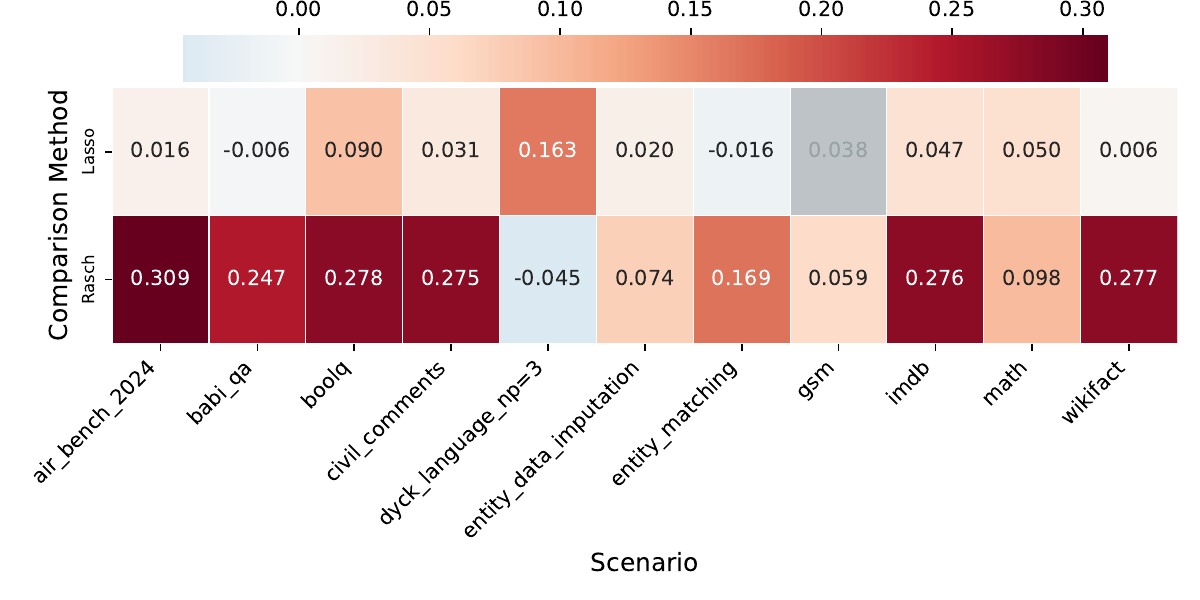}
                \caption{Heatmap of mean  differences in the 90th percentile of LLM-wise AUC  between     C-MIRT   and competing methods across scenarios.}

        \label{fig:compare_row}
    \end{figure}

In Figures \ref{fig:compare_col} and \ref{fig:compare_row}, each heatmap cell corresponds to a scenario–method pair. 
 The number in each cell is the average, across the five folds, of the difference in the 90th percentile of the AUC distribution between  C-MIRT and the competing method. A cell is shaded gray when its  paired t-test yields a $p$-value $>0.05$, indicating the difference is not statistically significant at the 5\% level. Among the remaining cells, red indicates that C-MIRT  performs better, whereas blue indicates that the competing method performs better.

Figure  \ref{fig:compare_col}   shows that for question-wise AUC,   C-MIRT consistently outperforms the Mean and Rasch baselines across different scenarios. In contrast, its comparison with the lasso method 
 is more scenario-dependent.  Figure \ref{fig:compare_row} shows that, however,  for LLM-wise AUC, C-MIRT remains competitive across the majority of scenarios. These results highlight the robustness of the proposed method. 
 
\section{Conclusions and future work}

This paper proposes to use  C-MIRT  for predicting evaluations of LLMs on unseen questions. 
We show that C-MIRT enhances     within-scenario prediction, whereas
cross-scenario generalizability depends on the source-target  scenarios and criterion. It opens an interesting future direction to further investigate new-scenario prediction for LLMs.

\section*{Impact Statement}

This paper studies a more efficient and interpretable approach to evaluating LLMs on unseen questions.
 A potential positive impact of this work is that it can reduce the cost of AI evaluation, both in annotation effort and computation, while still providing structured information about model strengths and weaknesses. However, the  method  also presents risks. Predicted performance may be mistaken for direct evidence of reliability, especially in high-stakes applications, and embedding-based item models may inherit biases from benchmark data or pretrained text representations. Our results also show that cross-scenario prediction is substantially harder than within-scenario prediction, so these methods should not be treated as a substitute for direct testing on representative data. We therefore view this work as a tool for supplementing evaluation pipelines, not replacing careful benchmark design, independent validation, or human oversight.

\bibliography{example_paper}
\bibliographystyle{icml2026}

\newpage

 
\appendix
\onecolumn
\section{Appendix}\label{sec:appendix}


\paragraph{Matrix representation and model identifiability.} 
For a fixed scenario \(s\), let \(Y^{(s)}\in\{0,1\}^{n\times p_s}\) be the response matrix, let
$
\boldsymbol{\alpha}^{(s)}=(\alpha_1^{(s)},\ldots,\alpha_n^{(s)})^\top\in\mathbb{R}^n,
$
let \(\mathbf{U}^{(s)}\in\mathbb{R}^{n\times r}\) collect the row vectors \({u_i^{(s)}}^\top\), and let \(\mathbf{V}^{(s)}\in\mathbb{R}^{p_s\times r}\) collect the transformed embeddings \(\phi^{(s)}(e_{j})^\top\). 

When the latent question factors are estimated freely rather than parameterized through embeddings, the model reduces to
\[
\boldsymbol{\Theta}^{(s)}=
\boldsymbol{\alpha}^{(s)}\mathbf{1}_{p_s}^\top+\mathbf{U}^{(s)}\mathbf{V}^{(s)\top}.
\]
In that case, to remove the usual rotational ambiguity, one may impose
\[
\mathbf{V}^{(s)\top}\mathbf{1}_{p_s}=\mathbf{0}_r,
\qquad
\mathbf{U}^{(s)\top}\mathbf{U}^{(s)}=\mathbf{V}^{(s)\top}\mathbf{V}^{(s)},
\]
under which the factorization is identifiable up to an orthogonal transformation~\citep{tang2026knowledge}. Specifically, if there exists another set of parameters $\{\bar{\boldsymbol{\alpha}}^{(s)} , \bar{\mathbf{U}}^{(s)} , \bar{\mathbf{V}}^{(s)}\} $ satisfying the same constraints such that $\boldsymbol{\alpha}^{(s)} \mathbf{1}_{p_s}^\top + \mathbf{U}^{(s)} \mathbf{V}^{(s)\top} = \bar{\boldsymbol{\alpha}} \mathbf{1}_{p_s}^\top + \bar{\mathbf{U}}^{(s)} \bar{\mathbf{V}}^{(s)\top}$, then$$\boldsymbol{\alpha}^{(s)} = \bar{\boldsymbol{\alpha}}^{(s)}, \quad \mathbf{U}^{(s)} = \bar{\mathbf{U}}^{(s)} \mathbf{O}, \quad \text{and} \quad \mathbf{V}^{(s)} = \bar{\mathbf{V}}^{(s)} \mathbf{O}, \quad \text{for some } \mathbf{O} \in \mathcal{O}(r),$$
where $\mathcal{O}(r)$ collects all orthonormal matrices in $\mathbb{R}^{r\times r}$.

\paragraph{Estimation procedure.}
Under the Bernoulli model with logistic link, we estimate the C-MIRT parameters using a two-stage procedure. In the first stage, for each scenario \(s\), we fit a low-rank logistic factorization model to obtain latent row and column representations \((\boldsymbol{\alpha}^{(s)}, \mathbf U^{(s)}, \mathbf V^{(s)})\). In the second stage, we learn the mapping \(\phi^{(s)}\) from contextual embeddings to the estimated question factors by supervised regression.

For a fixed scenario \(s\), the negative log-likelihood
under the Bernoulli model with logistic link is given by
\begin{align}
\mathcal{L}^{(s)}\bigl(\boldsymbol{\alpha}^{(s)},\mathbf U^{(s)},\mathbf V^{(s)}\bigr)
=
\sum_{i=1}^n\sum_{j\in\mathcal{T}_s}
\left[
\log\!\left(1+\exp\!\bigl(\theta_{ij}^{(s)}\bigr)\right)
-
y_{ij}^{(s)}\theta_{ij}^{(s)}
\right],
\label{eq:neg-like-matrix}
\end{align}
where $\mathcal{T}_s$, the index set, indicates those questions in the training data of scenario $s$.

To address the identifiability issue, we impose the centering constraint
\(
\mathbf V^{(s)\top}\mathbf 1_{p_s}=\mathbf 0_r,
\)
and we encourage balanced factorizations through the regularizer
\[
g\bigl(\mathbf U^{(s)},\mathbf V^{(s)}\bigr)
=
\left\|
\mathbf U^{(s)\top}\mathbf U^{(s)}
-
\mathbf V^{(s)\top}\mathbf V^{(s)}
\right\|_F^2.
\]
Accordingly, in stage 1 we solve the regularized optimization problem
\begin{align}
\min_{\boldsymbol{\alpha}^{(s)},\,\mathbf U^{(s)},\,\mathbf V^{(s)}}
\quad
\mathcal L_R^{(s)}\bigl(\boldsymbol{\alpha}^{(s)},\mathbf U^{(s)},\mathbf V^{(s)}\bigr)
:=
\mathcal L^{(s)}\bigl(\boldsymbol{\alpha}^{(s)},\mathbf U^{(s)},\mathbf V^{(s)}\bigr)
+
\frac{1}{4}
g\bigl(\mathbf U^{(s)},\mathbf V^{(s)}\bigr),
\label{eq:cmirt-reg-obj}
\end{align}
subject to
\[
\mathbf V^{(s)\top}\mathbf 1_{p_s}=\mathbf 0_r.
\]

We optimize \eqref{eq:cmirt-reg-obj} using projected gradient descent. At each iteration, we update \(\boldsymbol{\alpha}^{(s)}\), \(\mathbf U^{(s)}\), and \(\mathbf V^{(s)}\) by gradient descent, followed by a projection step that enforces \(\mathbf V^{(s)\top}\mathbf 1_{p_s}=\mathbf 0_r\).

After obtaining the stage-1 estimator
\[
\widehat{\mathbf{V}}^{(s)}=(\hat{v}_1^\top, \cdots, \hat{v}_j^\top, \cdots)_{j\in\mathcal{T}_s}\in\mathbb{R}^{p_s\times r},
\]
we proceed to stage 2 and estimate the feature map \(\phi^{(s)}\) by regressing \(\hat v_j^{(s)}\) on the corresponding contextual embedding \(e_j\). Specifically, we parameterize \(\phi^{(s)}\) by a multilayer perceptron and minimize the regression loss
\[
\mathcal L_{\mathrm{reg}}^{(s)}
=
\frac{1}{|\mathcal T_s^b|}
\sum_{j\in\mathcal T_s^b}
\left\|
\phi^{(s)}(e_j)-\hat v_j^{(s)}
\right\|_2^2
\]
over mini-batches \(\mathcal T_s^b\subseteq\mathcal T_s\). This yields an estimator \(\widehat{\phi}^{(s)}\), which can then be used to predict latent representations for unseen questions and hence out-of-sample response probabilities through
\[
\hat\theta_{ij}^{(s)}
=
\hat\alpha_i^{(s)}
+
\hat u_i^{(s)\top}\widehat{\phi}^{(s)}(e_j).
\]

\end{document}